\documentclass{article}
\usepackage{stywhispers,amsmath,epsfig}
\usepackage{url}
\usepackage{caption}
\usepackage{subcaption}
\usepackage{subfig} 
\usepackage{xcolor}
\usepackage{color}

\title{Combining multiscale features for classification of hyperspectral images: 
a sequence-based kernel approach}
\name{Yanwei Cui, Laetitia Chapel, S\'ebastien Lef{\`e}vre}
\address{Univ. Bretagne-Sud, UMR 6074, IRISA, F-56000 Vannes, France \\ \{yanwei.cui, laetitia.chapel, sebastien.lefevre\}@irisa.fr\\}
\begin{document}
%
\maketitle
\begin{abstract}
Nowadays, hyperspectral image classification widely copes with spatial information to improve accuracy. One of the most popular way to integrate such information is to extract hierarchical features from a multiscale segmentation.  In the classification context,  the extracted features are commonly concatenated into a long vector (also called stacked vector), on which is applied a conventional vector-based machine learning technique (\textit{e.g.} SVM with Gaussian kernel). In this paper, we rather propose to use a sequence structured kernel: the spectrum kernel. We show that the conventional stacked vector-based kernel is actually a special case of this kernel. Experiments conducted on various publicly available hyperspectral datasets illustrate the improvement of the proposed kernel \textit{w.r.t.} conventional ones using the same hierarchical spatial features. \footnote{8th IEEE GRSS Workshop on Hyperspectral Image and Signal Processing: Evolution in Remote Sensing (WHISPERS 2016),  UCLA in Los Angeles, California, U.S. \\ \url{http://www.ieee-whispers.com}}
 
\end{abstract}
\section{Introduction}
\label{sec:intro}

Integration of spatial information paves the way for improved accuracies in hyperspectral image classification \cite{fauvel2013advances}, as the use of spatial features extracted from image regions produces spatially smoother classification maps \cite{fauvel2012spatial}.
A common approach to extract such features is to rely on multiscale representations, \textit{e.g.} through (extracted) attribute profiles \cite{dalla2010extended} or hierarchical spatial features \cite{bruzzone2006multilevel}.
In this framework, features from multiple scales are extracted to model the context information around the pixels through different scales. Features computed at each scale are then concatenated into a unique (long) stacked vector. Such a vector is then used as input into a conventional classifier like SVM. Representative examples of this framework include \cite{bruzzone2006multilevel, lefevre14, huo2015semisupervised}. While defining kernels on stacked vectors is a simple and standard way to cope with hierarchical spatial features, it does not take into account the specific nature (\textit{i.e.} hierarchical) of the data.

Indeed, hierarchical spatial features extracted from hyperspectral images can rather be viewed as a sequence of data, for which structured kernels are commonly applied in other fields. Among the existing sequence structured kernels, the spectrum kernel based on subsequences of various lengths has been successfully applied in various domains, \textit{e.g.} biology for protein classification \cite{leslie2002spectrum,smola2003fast} or nature language processing for text classification \cite{lodhi2002text}. Its relevance for hyperspectral image classification remains to be demonstrated and is the main objective of this paper. Indeed, by applying the spectrum kernel onto hierarchical spatial features, we can explicitly take into account the  hierarchical relationships among regions from different scales. To do so, we construct kernels on various lengths of subsequences embedded in the whole set of hierarchical spatial features instead of modeling this set as a single stacked vector, the latter actually being  a particular case of the sequence kernel. Furthermore, we also propose an efficient algorithm to compute the spectrum kernel with all possible lengths, thus making realistic to apply such a kernel on hyperspectral images.

The paper is organized  as follows. We first briefly recall some background on hierarchical image representation (Sec.~\ref{sec:Hierarchical}). We then detail the concept of spectrum kernel (Sec.~\ref{sec:Spectrum}), and introduce an efficient algorithm for its computation. Evaluation of proposed method is detailed in Sec.~\ref{sec:Experiments}, before giving a conclusion and discussing future works.

\section{Hierarchical image representation}
\label{sec:Hierarchical}



\begin{figure}[htb]
\centering
\includegraphics[width=1.6cm,height=1.6cm]{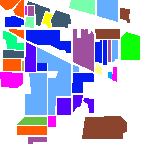}
\includegraphics[width=1.6cm,height=1.6cm]{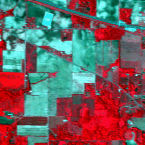}
\includegraphics[width=1.6cm,height=1.6cm]{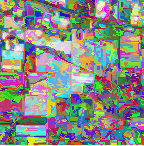}
\includegraphics[width=1.6cm,height=1.6cm]{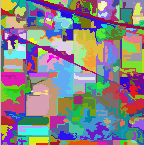}
\includegraphics[width=1.6cm,height=1.6cm]{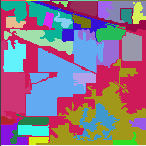}
\caption{The image representation of \texttt{Indian Pines} at different segmentation scales. From left to right: ground truth, false-color image,  fine level (2486 regions), intermediate level (278 regions),  coarse level (31 regions). }
\label{fig:hierarchcal}

\end{figure}

Hierarchical image representation describes the content of an image from fine to coarse level (as illustrated in Fig.~\ref{fig:hierarchcal}) through a tree structure,  where the nodes represent the image regions at different levels and the edges model the hierarchical relationships among those regions. Such representation is commonly used in the GEOgraphic-Object-Based Image Analysis (GEOBIA) framework \cite{Hay2008} and can be constructed with hierarchical segmentation algorithms, \textit{e.g.} HSeg \cite{tilton2010rhseg}. 
%

Let $n_1$ be a pixel of the image. Through hierarchical image representation, we write  $n_i$ the nested image regions at level $i= 2,...,p_\text{max}$, with region at lower levels always being included in higher levels \textit{i.e.} $n_1 \subseteq n_2 \ldots \subseteq n_{p_\text{max}}$. The context information of pixel $n_1$ can be then described by its ancestor regions $n_i$ at multiple levels $i= 2,...,p_\text{max}$. More specifically, one can define the context information as a sequence $S= \{n_1,...,n_{p_\text{max}}\}$ that encodes the evolution of the pixel $n_1$ through the different levels of the hierarchy. Each $n_{i}$ is described by a $D$-dimensional feature $\boldsymbol{x}_{i}$ that encodes the region characteristics \textit{e.g.} spectral information, size, shape, etc.

\section{ spectrum kernels}
\label{sec:Spectrum}

\subsection{Definition}
\label{ssec:Spectrum}

The spectrum kernel is an instance of kernels for structured data that allows the computation of similarities between contiguous subsequences of different lengths \cite{leslie2002spectrum,smola2003fast}. Originally designed for symbolic data, we propose here an adaptation to deal with hierarchical  representations equipped with numerical features.

\begin{figure}[htb]
\centering
\includegraphics[width = 0.35\textwidth]{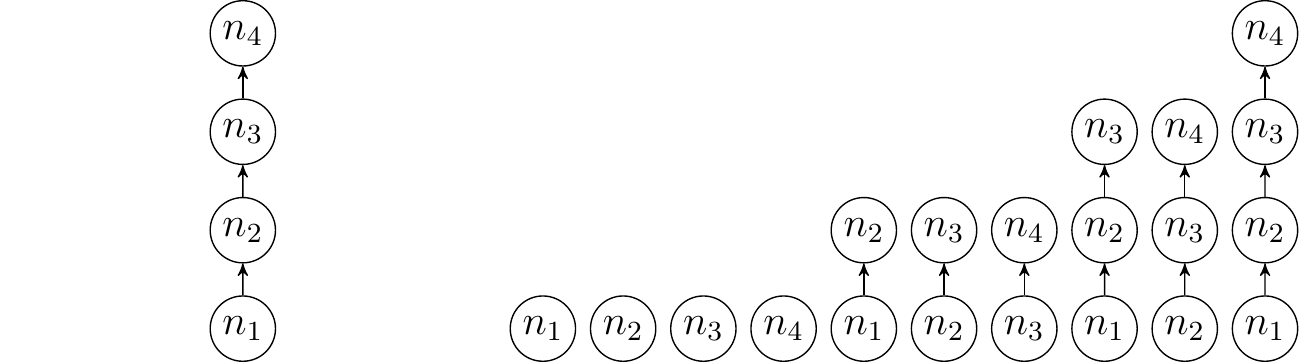}
\caption{A sequence  $S$ (left) and all its subsequences $s_p$ (right).}
\label{fig:sequence}
\end{figure}

Contiguous subsequences can be defined as $ s_{p} = (n_t, n_{t+1}$ $...,  n_{t+p})$, with $ t \geq 1, t+p \leq p_\text{max}$ and $p$ being the subsequence length. Fig.~\ref{fig:sequence} gives an example of a sequence and enumerates all its subsequences $s_p$. 

The spectrum kernel measures the similarity between two sequences $S, S'$ by summing up kernels computed on all their subsequences. Let $S_p = \{s_p \in S \mid |s_p| = p\}$ be the set of subsequences with a specific length $p$, the spectrum kernel can be written as:
\begin{equation} 
\begin{split}
K(S,S')& =  \sum_{p} \; \omega_{p} ~  K(S_p,S'_p) \; \\ 
 & = \sum_{p} \;  \omega_{p} \sum_{\substack{s_{p} \in S_p, s'_{p} \in S'_p }}   K(s_p,s'_p ) \;  ,
\label{spectrumSym}
\end{split}
\end{equation}
where the $p$-spectrum kernel $ K(S_p,S'_p)$ is computed between the set of subsequences of length $p$, and is further weighted by parameter $\omega_{p}$. In other words, it only allows the matching of subsequences with same length.  The kernel between two subsequences $K(s_p,s'_p )$ is defined as the product of atomic kernels computed on individual nodes $k(n_{t+i},n'_{t'+i})$, with $i$ denoting the position of nodes in the subsequence, following an ascending order $ 0 \leq i \leq p - 1$:
\begin{equation} 
K(s_{p},s'_{p})= \prod_{\substack{i = 0}}^{p-1} k(n_{t+i},n'_{t'+i} )\;  .
\label{eq:path}
\end{equation}

$K(S,S')$ in Eq.~\eqref{spectrumSym} suffers a common issue for structured kernels: the kernel value highly depends on the length of the sequences, as the number of compared substructures greatly increases with the length of sequence. One can mitigate this problem by normalizing the kernel as:
\begin{equation} 
	K^*(S,S')= \frac{K(S,S') }{ \sqrt{K(S,S)}\sqrt{K(S',S')}}. \;  
\end{equation}
In the sequel, we only use the normalized version $K^*$ of the kernel (written $K$ for the sake of simplicity).

\subsection{Weighting}
\label{ssec:weighting}

Several common weighting schemes \cite{smola2003fast} can be considered: 

\begin{itemize}
	\item $\omega_{p}= 1$ if $p = q$ and $\omega_{p}= 0$ otherwise,  yielding to a $q$-spectrum kernel considering only subsequence with a given length $q$: $K(S,S')=\sum\limits_{{s_{q} \in S_q, s'_{q} \in S'_q }} K(s_q,s'_q )$; 
	\item  $\omega_{p}= 1$ for all $p$, leading to a constant weighting with all lengths of subsequences;
	\item  $\omega_{p}= \lambda^p$ with $\lambda\in (0,1)$, an exponentially decaying weight  \textit{w.r.t.} the length of the subsequences.
\end{itemize}

It should be noted here that when using Gaussian kernel for the atomic kernel 
\begin{equation} 
k(n_{i},n_{i}') = \exp (-\gamma \lVert \boldsymbol{x}_{i}-\boldsymbol{x}'_{i}\rVert^2)\;  ,
\label{eq:rbf1}
\end{equation}
 the kernel computed on the stacked vector $\boldsymbol{z} = (\boldsymbol{x}_{1}, \cdots, \boldsymbol{x}_{p_\text{max}})$ comes down to the $p_\text{max}$-spectrum kernel:

\begin{equation} \label{eq:rbf2}
\begin{split}
  & K(s_{p_\text{max}},s'_{p_\text{max}}) =  \prod_{\substack{i = 1}}^{p_\text{max}} \exp (-\gamma \lVert \boldsymbol{x}_{i}-\boldsymbol{x}'_{i}\rVert^2)\;  = \\
& \exp \left(\sum_{\substack{i = 1}}^{p_\text{max}} (-\gamma \lVert \boldsymbol{x}_{i}-\boldsymbol{x}'_{i}\rVert^2)\right)\; = \exp (-\gamma \lVert \boldsymbol{z}-\boldsymbol{z}'\rVert^2)\;.
\end{split}
\end{equation}

\subsection{Kernel computation}
\label{ssec:computation}

We propose here an efficient computation scheme to iteratively compute all the $p$-spectrum kernels in a single run, yielding a complexity of $O(p_\text{max}p'_\text{max})$. 
The basic idea is to iteratively compute the kernel on subsequences $s_{p}$ and $s'_p$ using previously computed kernels on subsequences of length $(p-1)$. The atomic kernel $k(n_{i}, n'_{i'})$ thus needs to be computed only once, avoiding redundant computing.

We define a three-dimensional matrix $M$ of size $p_\text{max} \times p'_\text{max} \times \min(p_\text{max}, p'_\text{max})$, where each element $M_{i,i',p}$ is defined as:
\begin{equation} 
M_{i,i',p} = k(n_i,n'_{i'}) ( M_{{i-1},{i'-1} ,p-1} ) \;  .
\label{eq:computation}
\end{equation} 
where $M_{0,0,0} = M_{0,i',0} = M _{i,0,0} = 1$ by convention.
The kernel value for the $p$-spectrum kernel is then computed as the sum of all the matrix elements for a given $p$:
 \begin{equation} 
 K(S_p,S'_p) = \sum_{i,i'=1}^{p_{\text{max}},p'_{\text{max}}}M_{i,i',p} \;  .
 \end{equation}  

\section{Experiments}
\label{sec:Experiments}

\subsection{Datasets and design of experiments}

We conduct experiments on 6 standard hyperspectral image datasets: \texttt{Indian Pines}, \texttt{Salinas}, \texttt{Pavia Centre} and \texttt{University}, \texttt{Kennedy space center (KSC)} and \texttt{Botswana}, 
considering a \textit{one-against-one} SVM classifier (using the Java implementation of LibSVM \cite{chang2011}).

We use Gaussian kernel as the atomic kernel $ k(\cdot,\cdot)$. 
Free parameters are determined by 5-fold cross-validation over potential values: the bandwidth $\gamma$ (Eq.~\eqref{eq:rbf1}) and the SVM regularization parameter $C$.  We also cross-validate the different weighting scheme parameters: $q \in \{1, \ldots, p_\text{max}\}$ for the $q$-spectrum kernel and  $\lambda\in (0,1)$ for the decaying factor.

\subsection{Results and analysis}

We randomly pick $n=\{10,25,50\}$ samples per class from available ground truth for training, and the rest for testing.  In the case of small number of pixels per class in \texttt{Indian Pines} dataset (total sample size for a class less than $2n$), we use half of samples for training.

Hierarchical image representations are generated with HSeg \cite{tilton2010rhseg} by increasing the region dissimilarity criterion $\alpha$. Parameter $\alpha$ is empirically chosen: $\alpha=[2^{-2},2^{-1},...,2^{8}]$, leading to a tree that covers the whole scales from fine to coarse (top levels of whole image are discarded as they do not provide any additional information). Hierarchical levels $\alpha = \{2^2,2^4,2^6\}$ of \texttt{Indian Pines} are shown in Fig.~\ref{fig:hierarchcal} as the fine, intermediate, coarse level for illustration. Features $\boldsymbol{x}_{i}$ that describe each region are set as the average spectral information of the pixels that compose the  region.

\subsubsection{Comparison with state-of-the-art algorithms}

We compare our sequence-based kernel with state-of-the-art algorithms that take into account the spatial information relying on multiscale representation of an image: i) spatial-spectral kernel \cite{fauvel2012spatial} that uses area filtering to obtain the spatial features (the filtering size is fixed so as to lead to the best accuracy); ii) attribute profile \cite{dalla2010extended}, using 4 first principal components with automatic level selection for the area attribute and standard deviation attribute as detailed in \cite{ghamisi2014automatic}; iii) hierarchical features stored on a stacked vector \cite{bruzzone2006multilevel,lefevre14,huo2015semisupervised}. 
For comparison purposes, we also report the pixel-based classification overall accuracies.
All results are obtained by averaging the performances over 10 runs of (identical among the algorithms) randomly chosen training and test sets.

First of all, in Tab.~\ref{tab:results}, we can see that the overall accuracies are highly improved when spatial information is included. Using hierarchical features computed over a tree (stacked vector or any version of the spectrum kernel) yields competitive results compared with state-of-the-art methods. By applying the proposed spectrum kernel on the hierarchical  features rather than a kernel on a stacked vector, the results are further improved: best results for \texttt{Indian Pines}, \texttt{Salinas}, \texttt{Pavia Centre}, \texttt{KSC} and \texttt{Botswana} datasets are obtained with a spectrum kernel. We can observe that attribute profiles perform better for \texttt{Pavia University}. This might be due to the kind of hierarchical representation used, \textit{i.e.} min and max-trees in the case of attribute profiles instead of HSeg in our case. Besides, the popularity of these profiles as well as the \texttt{Pavia} dataset result in optimizations of the scale parameters for years. However, the proposed spectrum kernel is not limited at all to the HSeg representation, and it is thus possible to apply it to min- and max-trees and attribute features. This will be explored in future studies.

\subsubsection{Impact of the weighting scheme}
\begin{figure*}[!htb]
	\centering
	
\begin{subfigure}[t]{0.32\textwidth}
	\includegraphics[width=1\textwidth]{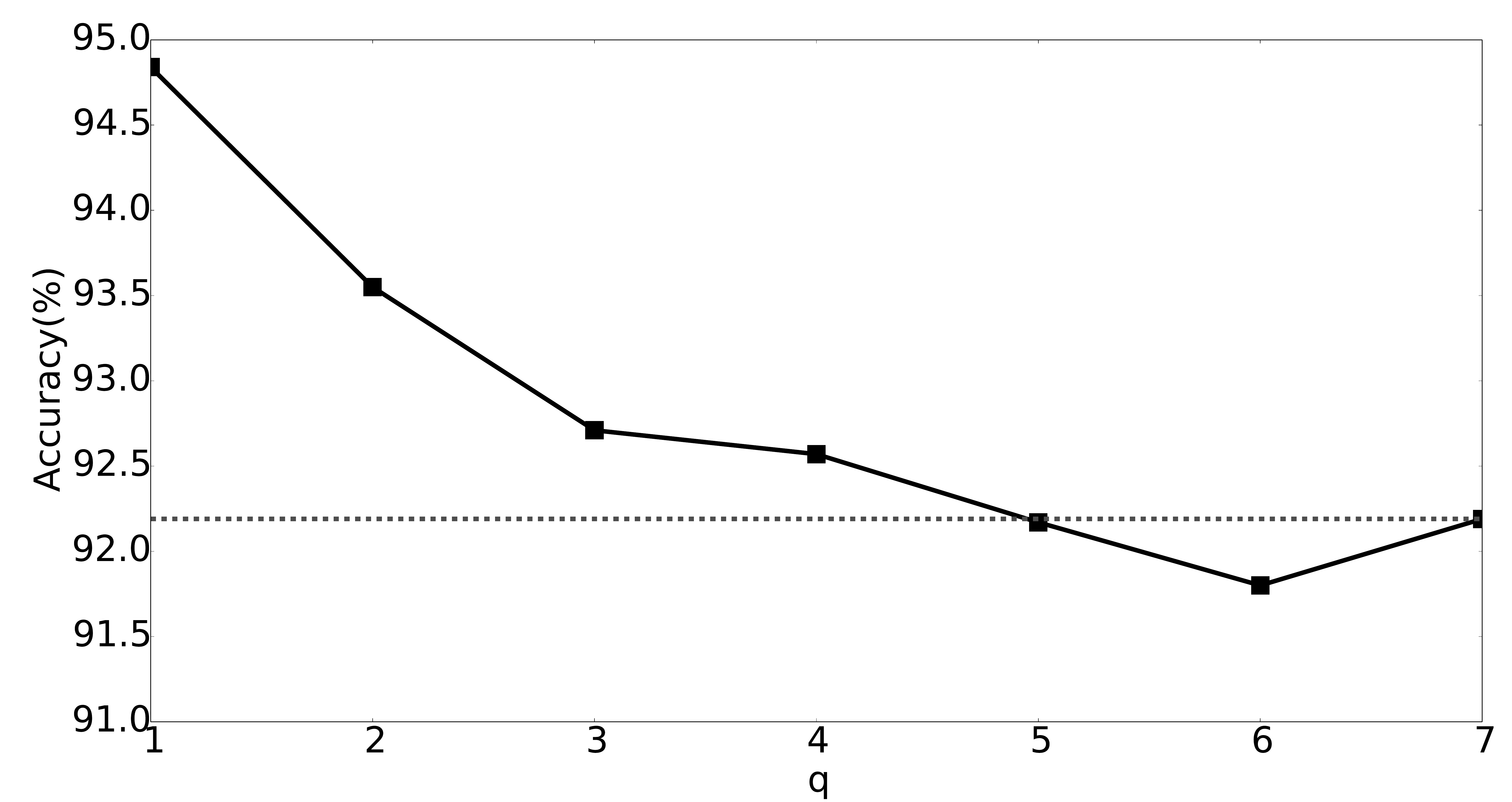}
\caption{\texttt{Indian Pines}}
\end{subfigure}
\begin{subfigure}[t]{0.32\textwidth}
\includegraphics[width=1\textwidth]{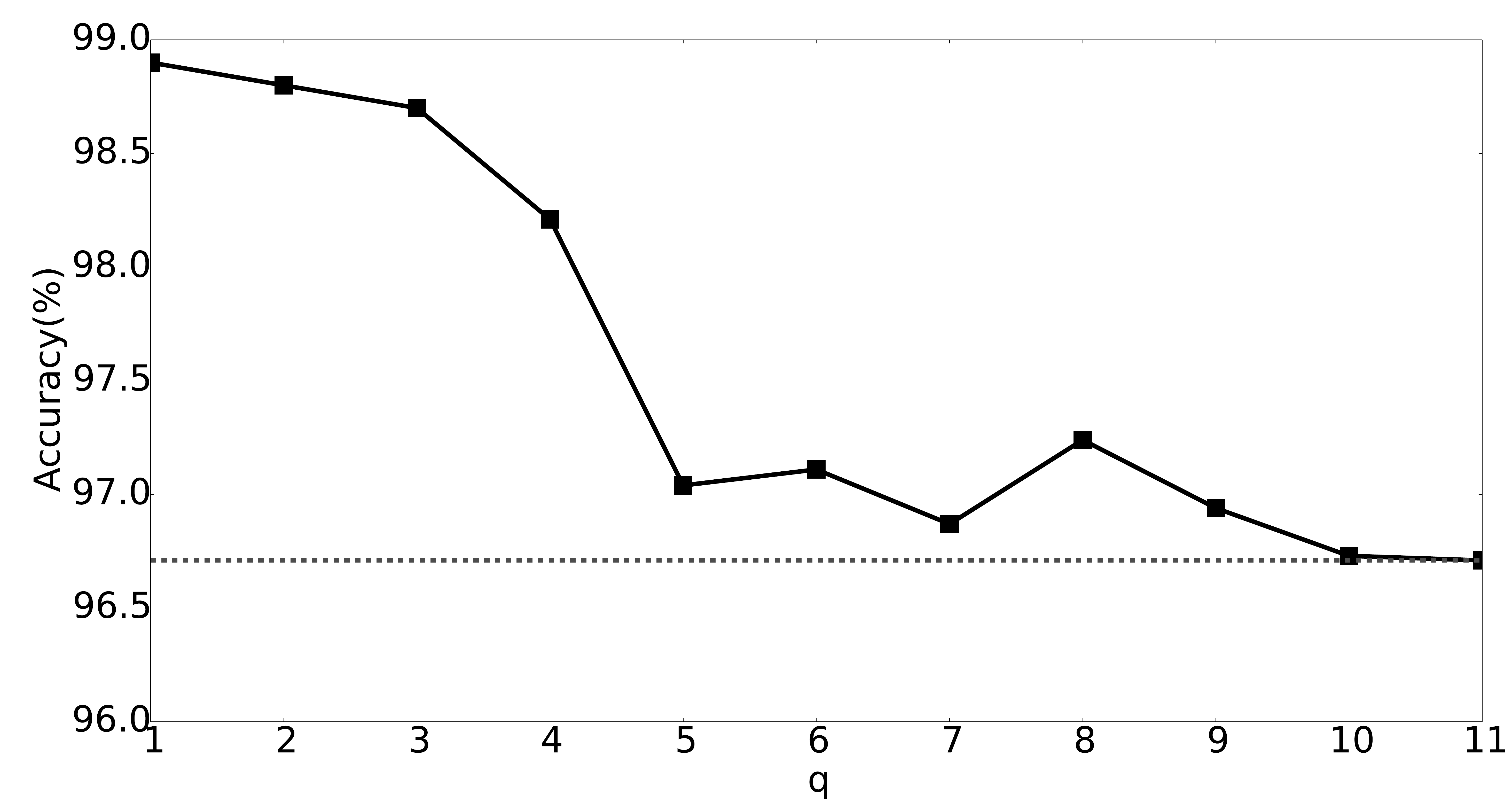}
\caption{\texttt{Salinas}}
\end{subfigure}
\begin{subfigure}[t]{0.32\textwidth}
	\includegraphics[width=1\textwidth]{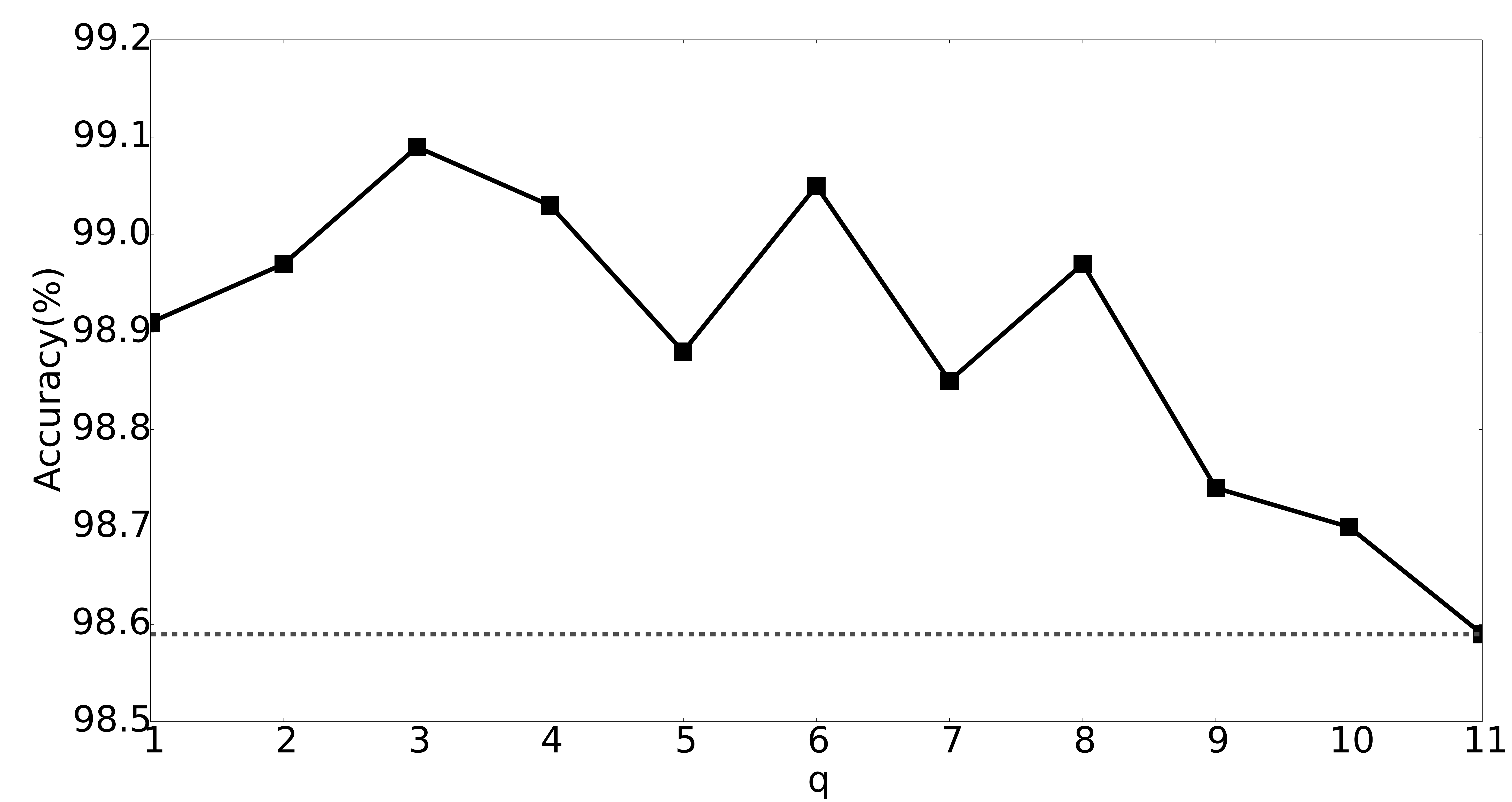}
\caption{\texttt{Pavia Centre}}
\end{subfigure}

\begin{subfigure}[t]{0.32\textwidth}
	\includegraphics[width=1\textwidth]{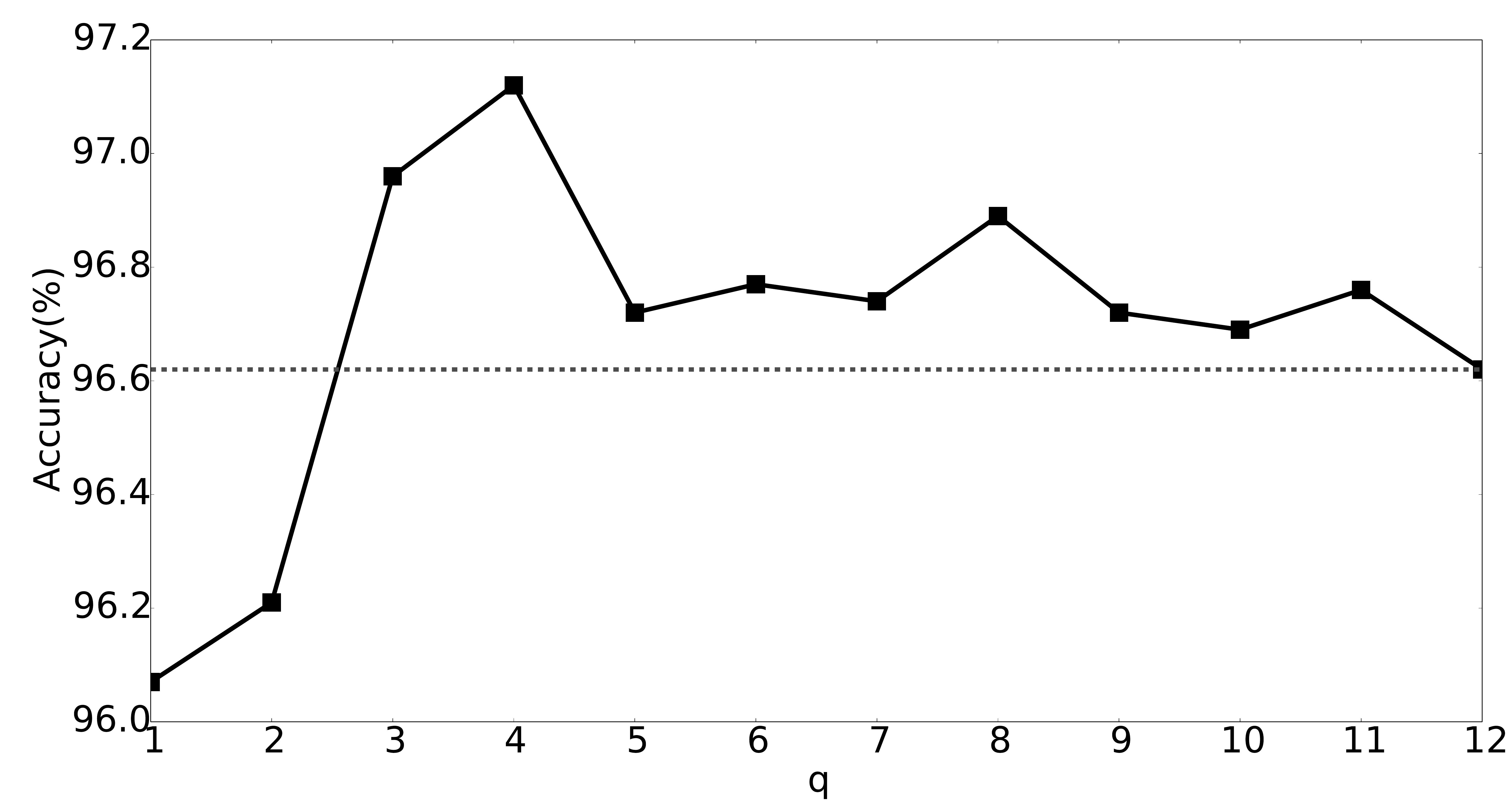}
\caption{\texttt{Pavia University}}
\end{subfigure}
\begin{subfigure}[t]{0.32\textwidth}
	\includegraphics[width=1\textwidth]{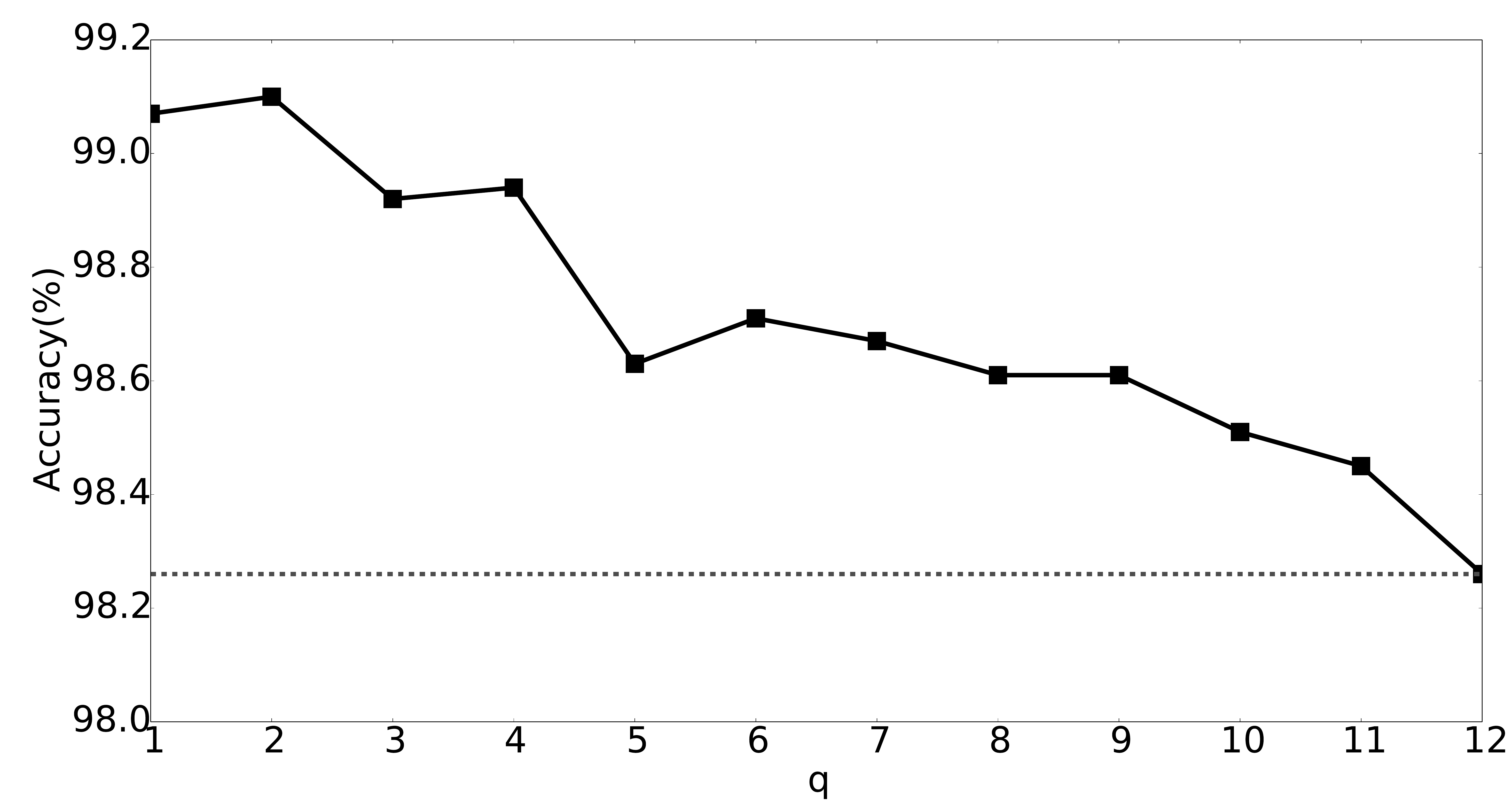}
\caption{\texttt{KSC}}
\end{subfigure}
\begin{subfigure}[t]{0.32\textwidth}
	\includegraphics[width=1\textwidth]{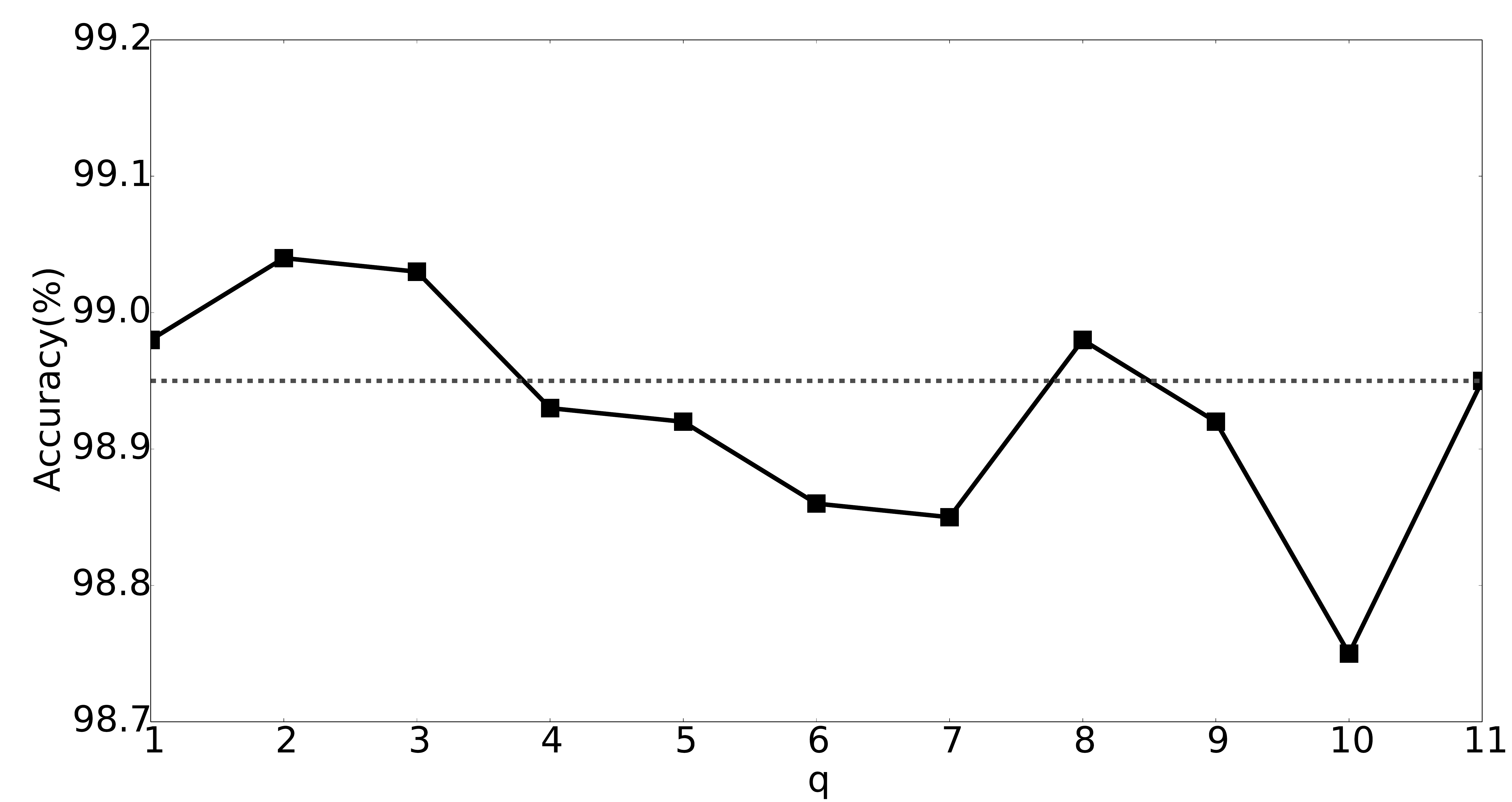}
\caption{\texttt{Botswana}}
\end{subfigure}

\caption{The overall accuracies of $q$-spectrum kernel with different lengths $q$ using $n = 50$ training samples per class. Results for the stacked vector with Gaussian kernel are shown in dashed line.}
\label{fig:spectrum}

\end{figure*}

We study the impact of the different weighting schemes introduced in Sec.~\ref{ssec:weighting}. Fig.~\ref{fig:spectrum} shows that the stacked vector ($q = p_\text{max}$) does not lead to the best performances, and that the best scale $q$ can not be determined beforehand as it depends on the dataset. For most setups, combination of different scales (constant weighting or decaying factor) allows the improvement of the accuracies. However, the best weighting scheme again depends on the considered dataset, and this calls for a more extensive study of weighting strategies.



\begin{table*}[htb]
	\footnotesize
	\centering

	\caption{Mean (and standard deviation) of overall accuracies (OA) computed over 10 repetitions using $n$ training samples per class for 6 hyperspectral image datasets. $c$ stands for constant weighting, $q$ for the $q$-spectrum kernel and $\lambda$ for the decaying weight. Best results are boldfaced.}
	\begin{tabular}{|c|c||c|c|c||c|c|c|}
		\hline
		& \multicolumn{7}{c|}{\texttt{Indian Pines}}  \\ 
		\hline
		$n$ & pixel only & Spatial-spectral   & Attribute profile & Stacked vector  & Spectrum kernel-$c$ & Spectrum kernel-$q$ & Spectrum kernel-$\lambda$  \\
		\hline
		10 & 54.89 (2.10) & 72.03 (2.52)& 64.37 (2.87) & 73.21 (2.60)& 78.70 (4.88)  &  80.19 (4.48) & \textbf{80.19 (3.40)}\\

		25 & 66.04 (1.59) & 84.02 (1.31) & 76.71 (2.60) & 84.90 (2.42)& 89.16 (2.89) & \textbf {91.36 (1.57)} & 89.46 (3.61)\\

		50 & 72.99 (0.10) & 90.82 (2.07)& 84.57 (1.45) & 92.19 (0.86)& 94.12 (1.18) & \textbf{94.76 (1.09)} & 94.48 (1.20) \\
		\hline
	\end{tabular}
		\begin{tabular}{|c|c||c|c|c||c|c|c|}
		\hline
		& \multicolumn{7}{c|}{\texttt{Salinas}}  \\ 
		\hline
		$n$ & pixel only & Spatial-spectral     & Attribute profile & Stacked vector  & Spectrum kernel-$c$ & Spectrum kernel-$q$ & Spectrum kernel-$\lambda$ \\
		\hline
		10 & 83.87 (1.96) & 87.72 (1.88)    & 91.89 (1.73) & 89.17 (2.95) & \textbf{93.18 (1.70)}  & 91.16 (2.65)  & 91.44 (2.71) \\

		25 & 88.13 (1.22) & 92.93 (0.98)    & 95.99 (1.11) & 94.86 (1.58) & \textbf{97.28 (1.62)}  & 97.04 (1.28)   & 97.02 (1.57) \\

		50 & 88.86 (1.22) & 94.34 (0.81)    & 97.39 (0.45) & 96.71 (0.70) & 98.51 (0.89)  & \textbf{98.81 (0.70)}   &  97.93 (1.22)\\
		\hline
	\end{tabular}
		\begin{tabular}{|c|c||c|c|c||c|c|c|}
		\hline
		& \multicolumn{7}{c|}{\texttt{Pavia Centre}}  \\ 
		\hline
		$n$ & pixel only & Spatial-spectral & Attribute profile & Stacked vector  & Spectrum kernel-$c$ & Spectrum kernel-$q$ & Spectrum kernel-$\lambda$  \\
		\hline
		10 & 93.37 (3.59) & 95.69 (0.73)  & 96.03 (0.91)  & 95.94 (1.01) &  96.14 (1.61) & 96.56 (1.09) & \textbf{96.71 (0.97)} \\

		25 & 96.13 (0.48) & 96.99 (0.48) &  97.59 (0.27) & 97.85 (0.53) & 97.93 (0.55)   & \textbf{97.96 (0.59)} & 97.93 (0.57) \\

		50 & 96.98 (0.52) & 98.10 (0.34) &  98.59 (0.24)  & 98.59 (0.48)&  98.83 (0.39)  &  98.92 (0.37) & \textbf{99.04 (0.31)}\\
		\hline
	\end{tabular}	
		\begin{tabular}{|c|c||c|c|c||c|c|c|}
		\hline
		& \multicolumn{7}{c|}{\texttt{Pavia University}}  \\ 
		\hline
		$n$ & pixel only & Spatial-spectral     & Attribute profile & Stacked vector  & Spectrum kernel-$c$ & Spectrum kernel-$q$ & Spectrum kernel-$\lambda$   \\
		\hline
		10& 69.00 (5.68) & 76.74 (5.26)    & \textbf{88.69 (4.06)} & 83.30 (3.75) & 84.34 (5.14) & 84.43 (6.13) & 85.10 (6.65) \\

		25& 79.81 (1.42) & 87.92 (3.36)    & \textbf{95.17 (1.84)} & 92.95 (3.29) & 93.70 (2.56) &  93.98 (1.91) & 93.98 (2.22) \\

		50& 84.72 (1.32) & 93.27 (1.29)    & \textbf{97.52 (0.86)} & 96.62 (1.06) & 97.20 (0.97) & 96.76 (1.11) &  96.66 (1.84) \\
		\hline
	\end{tabular}

		\begin{tabular}{|c|c||c|c|c||c|c|c|}
		\hline
		& \multicolumn{7}{c|}{\texttt{KSC}} \\ 
		\hline
		$n$ & pixel only & Spatial-spectral     & Attribute profile & Stacked vector & Spectrum kernel-$c$ & Spectrum kernel-$q$ & Spectrum kernel-$\lambda$  \\
		\hline
		10 & 86.56 (1.33) & 90.96(2.12)    & 90.61 (0.63) & 92.75 (1.71) & 93.98 (1.29) &  \textbf{94.18 (0.87)} & 94.01 (1.15)\\

		25 & 91.27 (0.84) & 97.16 (0.16)    & 95.53 (0.71) & 97.32 (0.45)  & \textbf{97.85 (0.63)} & 97.45 (0.86) & 97.82 (0.66)  \\

		50 & 93.67 (0.58) & 98.46 (0.29)    & 97.41 (0.49) & 98.26 (0.37) & 99.13 (0.34) & 99.00 (0.40)  & \textbf{99.15 (0.23)} \\
		\hline
	\end{tabular}
	
		\begin{tabular}{|c|c||c|c|c||c|c|c|}
		\hline
		& \multicolumn{7}{c|}{\texttt{Botswana}}  \\ 
		\hline
		$n$ & pixel only & Spatial-spectral & Attribute profile & Stacked vector  & Spectrum kernel-$c$ & Spectrum kernel-$q$ & Spectrum kernel-$\lambda$  \\
		\hline
		10 & 87.72 (2.42) & 92.62 (1.40) & 92.17 (1.32)  & 94.16 (1.41) &  \textbf{94.66 (1.62)}  & 94.59 (1.69) & 94.63 (1.54) \\

		25 & 91.89 (0.67) & 96.65 (0.69) &  95.35 (0.91) & 97.71 (0.72) & \textbf{97.99 (0.48)}   & 97.79 (0.55) &  97.90 (0.79)\\

		50 & 94.03 (0.60) &  97.74 (0.52) & 96.83 (0.64)  & 98.95 (0.53)&  \textbf{99.10 (0.50)}  &  98.97 (0.42) & 98.99 (0.45)  \\
		\hline
	\end{tabular}	
	\label{tab:results}
\end{table*}

\section{Conclusion}

In this paper, we propose to use the spectrum kernel for applying machine learning on hierarchical features for hyperspectral image classification.  The proposed kernel considers the hierarchical features as a sequence of data and exploits the hierarchical relationship among regions at multiple scales by constructing kernels on various lengths of subsequences. The method exhibits better performances than state-of-the-art algorithms for all but one tested dataset. We also show that combining different scales allows the improvement of the accuracies, but the way to combine them should be further explored. The use of optimal weights thanks to the multiple kernel learning framework \cite{gonen2011multiple} is the next step of our work.

\bibliographystyle{IEEEbib}
\bibliography{refs}

\end{document}